# "Bridging the Gap: An Intermediate Language for Enhanced and Cost-Effective Grapheme-to-Phoneme Conversion with Homographs with Multiple Pronunciations Disambiguation"

Abbas Bertina[1]   Shahab Beirami[1]   Hossein Biniazian[1]   Elham Esmaeilnia[1]   Soheil Shahi[1]
Mahdi Pirnia[1]

[1]Dept. of Artificial Intelligence, Bertix AI

*a.bertina@bertix.ai, sh.beirami@bertix.ai, h.biniazian@bertix.ai, e.esmaeilinia@bertix.ai,
s.shahi@bertix.ai, m.pirnia@bertix.ai*

*Abstract*—Grapheme-to-phoneme (G2P) conversion for Persian presents unique challenges due to its complex phonological features, particularly homographs and Ezafe, which exist in formal and informal language contexts. This paper introduces an intermediate language specifically designed for Persian language processing that addresses these challenges through a multi-faceted approach. Our methodology combines two key components: Large Language Model (LLM) prompting techniques and a specialized sequence-to-sequence machine transliteration architecture. We developed and implemented a systematic approach for constructing a comprehensive lexical database for homographs with multiple pronunciations disambiguation often termed polyphones, utilizing formal concept analysis for semantic differentiation. We train our model using two distinct datasets: the LLM-generated dataset for formal and informal Persian and the B-Plus podcasts for informal language variants. The experimental results demonstrate superior performance compared to existing state-of-the-art approaches, particularly in handling the complexities of Persian phoneme conversion. Our model significantly improves Phoneme Error Rate (PER) metrics, establishing a new benchmark for Persian G2P conversion accuracy. This work contributes to the growing research in low-resource language processing and provides a robust solution for Persian text-to-speech systems and demonstrating its applicability beyond Persian. Specifically, the approach can extend to languages with rich homographic phenomena such as Chinese and Arabic.

## 1. Introduction

Grapheme-to-phoneme (G2P) conversion represents a significant challenge in speech processing and natural language understanding, particularly for languages with complex phonological systems. As a morphologically rich and phonetically ambiguous language, Persian presents unique difficulties due to the absence of diacritics in standard writing, making pronunciation and meaning highly context-dependent. Furthermore, Persian suffers from homographs with multiple pronunciations and meanings ambiguity also referred to as *polyphones*, where words with identical spelling can have multiple pronunciations and meanings, as well as the Ezafe phenomenon, which involves phonetic markers (both "کسره-اضافه" and "ه-کسره"), with the latter not being explicitly represented in the script. Additionally, the stark contrast between formal and informal Persian further complicates computational phonological analysis. To address these challenges, we propose an **intermediate language representation** that can be applied to Persian and other complex phonological languages, facilitating a more structured and adaptable G2P conversion process. Our approach integrates the strengths of neural machine transliteration (NMT) with sophisticated data augmentation techniques, leveraging the machine transliteration (MT) framework specifically optimized for Persian G2P conversion. By introducing an intermediate representation, we aim to enhance the generalizability of our model, making it adaptable to other languages with similar phonetic complexities. Our key contributions include:

- A novel framework for dataset generation via adaptive prompt-based learning mechanisms
- A systematic data augmentation methodology for training set expansion with preserved linguistic fidelity
- Domain-specific tokenization architecture for intermediate language processing
- A robust G2P conversion system that effectively handles Persian-specific challenges, including Ezafe prediction and homograph disambiguation, both for formal and informal language

Our approach demonstrates significant improvements over existing methods, achieving a BLEU score of 94.6 and effectively handling complex phonological phenomena. It also showcases our solution's robustness. This success is particularly noteworthy given the challenges of Persian language processing and the limited availability of labeled data in this domain.

The remainder of this paper is organized as follows: Section 2 reviews related works, Section 3 details our methodology, including dataset construction for Persian G2P conversion and the model architecture, Section 4 describes our experimental setup and evaluation metrics, Section 5 presents our results and analysis, and Section 6 concludes our work. Finally, section 7 proposes future research directions.

## 2. Related Work

Recent advances in G2P conversion have demonstrated various innovative approaches and methodologies. The Persian phonemizer represents a significant advancement in Persian language processing, offering an automated solution for converting Persian text to the International Phonetic Alphabet (IPA) [22] notation through an integrated approach combining dictionary-based lookups, machine learning, and linguistic rules. The system's architecture intelligently handles one of Persian's main challenges "words with multiple pronunciations" by utilizing a multi-step process that includes POS tagging, dependency parsing, and a neural network-based grapheme-to-phoneme model. However, the tool faces notable limitations, including restricted dictionary coverage, potential accuracy issues with modern terminology, and computational overhead from its complex processing pipeline [18]. PersianG2P represents another approach to Persian grapheme-to-phoneme conversion, employing a hybrid system of dictionary lookups and neural network predictions to handle Persian text's unique challenges, including unwritten vowels and context-dependent pronunciations. The tool offers two dictionary options (1,867 and 48,000 words) and utilizes Hazm [20] normalization for text processing, demonstrating improved accuracy compared to existing tools like Epitran[1]. However, the system's limitations include restricted dictionary coverage for modern terms and potential inconsistencies in handling complex context-dependent pronunciations, highlighting the ongoing challenges in Persian language processing [19].

Rabiee [21] developed Persian_g2p, a sequence-to-sequence model for Persian grapheme-to-phoneme conversion designed for text-to-speech preprocessing. Their system combines the Tihu pronunciation dictionary[2] With a GRU-RNN seq2seq model to handle out-of-vocabulary (OOV) words, achieving a Phoneme Error Rate (PER) of 3.9% on their test set. Despite utilizing Hazm for text normalization and num2fawords for number conversion, the system shows limitations in processing informal words, slang expressions, and broken language patterns common in contemporary Persian usage. Researchers have evaluated the performance of Large Language Models (LLMs) in G2P conversion for the Persian language, introducing innovative prompting and post-processing methods that enhance LLM outputs without additional training or labeled data [1]. Their work resulted in significant contributions, including the creation of "Kaamel-Dict,"[2] the Persian G2P dictionary containing over 120,000 entries, and "Sentence-Bench," the first sentence-level benchmarking dataset [3].

Alternative approaches to language modeling have shown promise through tokenization-free, phoneme, and grapheme-based language models [4]. These studies demonstrated that small models based on the Llama architecture[3] could achieve strong linguistic performance using character-level vocabularies, though limitations related to data choice, G2P tool selection, and architectural decisions were noted.

Resource constraints in G2P conversion have been addressed through innovative methodologies, such as the two-step approach for French pronunciation learning [5]. This method mitigates the lack of extensive labeled data by decomposing the complex pronunciation task into two sub-tasks. The Fish-Speech framework introduced a serial fast-slow Dual Autoregressive architecture, enhancing stability in Grouped Finite Scalar Vector Quantization (GFSQ) for sequence generation [6].

Significant advancement in contextual G2P conversion emerged with the introduction of in-context knowledge retrieval (ICKR) capabilities using GPT-4 [7], achieving a 95.7% accuracy rate and a 4.9% weighted average phoneme error rate on the LibriG2P dataset. The Interleaved Speech-Text Language Model (IST-LM) introduced a novel approach to streaming zero-shot Text-to-Speech synthesis [8], eliminating the need for traditional duration prediction and grapheme-to-phoneme alignment.

Recent developments in language-specific applications include a novel end-to-end framework for Chinese polyphone disambiguation [9] and BreezyVoice [10], a specialized TTS system for Taiwanese Mandarin. These systems demonstrate the effectiveness of combining pre-trained models with neural network classifiers and addressing specific linguistic challenges through innovative approaches.

---

[1] https://github.com/dmort27/epitran

[2] https://github.com/tihu-nlp/tihudict

[3] https://ai.meta.com/blog/meta-llama-3/

## 3. Methodology

This section introduces our methodology, including dataset construction for Persian G2P conversion, data collection, proposed intermediate language, and model architecture. The architectural overview of the proposed system is depicted in Figure 1, demonstrating the
pipeline from data collection to phoneme conversion.

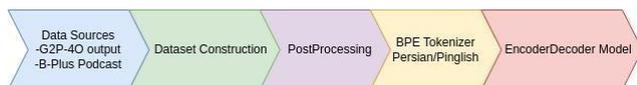

**Fig.1.** Schematic overview of the proposed G2P conversion methodology.

### 3.1 Dataset Construction for Persian G2P Conversion

Constructing a robust dataset for Persian G2P conversion was a pivotal step in our research, addressing the lack of annotated data and the specific challenges posed by homographs and Ezafe. This paper defines homographs as words that share the exact spelling but have different pronunciations and meanings also referred to as *polyphones*. For example, the Persian word "شیر" is not considered a homograph in our study, despite having multiple meanings, because its pronunciation remains the same across different contexts. However, the word "بر" qualifies as a homograph, as it has three distinct meanings and pronunciations: (1) as a noun referring to a type of animal (tiger), (2) as an imperative verb meaning "carry," and (3) as another imperative verb meaning "cut," each with a different pronunciation. Since Persian script does not include diacritical marks in standard writing, the meaning and pronunciation of words are inferred from the context of the sentence. In practice, language speakers memorize these pronunciations and meanings through usage. Here, we detail our methodology, and the architectural overview of the proposed dataset construction is depicted in Figure 2.

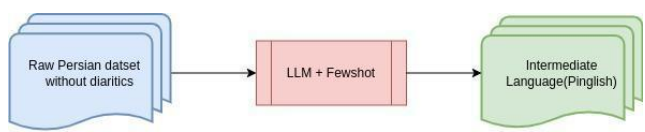

**Fig.2.** Schematic overview of the proposed dataset construction.

#### 3.1.1 Data Collection
The Persian language presents a distinctive characteristic where formal and informal variants of words differ in their written form and pronunciation, posing a significant challenge for Grapheme-to-Phoneme (G2P) models. Our method effectively addresses this dual variation between formal and informal word pairs. Table 1 demonstrates this phenomenon through representative examples of formal and informal word pairs, showing their distinct written forms and corresponding pronunciations. So we collect both formal and informal data for our proposed method:

- Formal Persian Data: We employed a generative approach using the advanced capabilities of GPT-4o[11]. By providing the model with 20 distinct topics, we prompted it to generate text in formal and informal Persian, ensuring a diverse representation of language usage. This method yielded 1 million raw Persian sentences without diacritics.
- Informal Persian Data: To capture colloquial expressions and everyday language, we sourced subtitles from the B-Plus podcast on YouTube [16]. This gave us a rich corpus of informal Persian text, further expanding our dataset.

**Table 1.** Formal and Informal Word Variations in Persian.

| | Informal | Formal |
|---|---|---|
| Persian infrmal-formal example | اونجا | آنجا |
| | کتابا | کتاب ها |
| | خوبه | خوب است |

#### 3.1.2 Intermediate Language Development
Given the absence of an existing dataset tailored for our intermediate language, we adopted a multi-step process:

**Romanized Text Generation:** Using a few-shot learning approach with an LLM, we generated a Romanized version of Persian called "**Pinglish**." This served as our intermediate language, designed to facilitate phoneme-to-character mapping. The LLM was trained on our few-shot examples to produce coherent Pinglish text.

**Post-processing:** To ensure consistency and reduce error rates, we applied a rule-based method to map each Persian phoneme to a single character in the intermediate language. This post-processing step was crucial to avoid the ambiguity often found in the IPA, where multiple characters might represent the same phoneme. By ensuring that each phoneme in Persian had a unique character representation in our intermediate language, we established a one-to-one correspondence, enhancing readability and reducing potential conversion errors.

#### 3.1.3 Dataset Creation

After post-processing, we amassed a dataset of 64,000 entries, each containing:

- The original Persian text.
- Its corresponding Pinglish representation.

#### 3.1.3.1 This dataset was meticulously curated to

- Ensure that a single character in our intermediate language could accurately represent every Persian phoneme.
- Provide a comprehensive foundation for training our G2P model, enabling it to handle formal and informal Persian accurately.

### 3.1.4 Advantages of the Intermediate Language

Prior studies on Persian G2P conversion have predominantly relied on rule-based approaches using the IPA and lexicon-driven mappings. While these methods offer phonetic accuracy in controlled settings, they face several limitations. Notably, they struggle with homograph disambiguation and Ezafe detection, particularly in the absence of broader contextual information. Additionally, their performance significantly degrades in the presence of morphological variations such as prefixes and suffixes. In our work, we deliberately avoided using IPA for several key reasons. First, IPA representations are composed of multiple characters per phoneme, which increases the complexity of the input and hampers the learning process in sequence-to-sequence models, such as those used in TTS systems. This representation also poses challenges for readability and manual verification due to its linguistic specificity and lack of intuitive clarity. Second, IPA exhibits a high degree of variation in symbol usage for the same phoneme, introducing redundancy and ambiguity that negatively impact the convergence and generalization capacity of neural models. In contrast, our proposed intermediate language is designed to be compact, unambiguous, and easily learnable. It addresses the specific requirements of Persian phonology while offering scalability to other languages. The development of this intermediate language addressed the specific needs of Persian G2P conversion and laid the groundwork for its application in other languages. Its design allows for:

- Consistency: Each phoneme has a unique character representation, reducing ambiguity.
- Scalability: The methodology can be adapted to other languages, not just Persian, making it a versatile tool for linguistic processing.

Through this systematic approach to dataset construction, we have created a valuable resource for Persian G2P conversion research. We have significantly reduced the complexity and cost of phoneme annotation and provided a robust solution for homograph disambiguation and Ezafe detection.

In our proposed intermediate language system, we established a one-to-one mapping between Persian phonemes and Latin characters, implementing specific modifications to standard Pinglish transliteration. To maintain phonetic consistency and eliminate ambiguous character mappings, we excluded the characters {c, u, q, w, x} from the character set. This exclusion was necessary as these characters either lack direct phonemic equivalents in Persian or create ambiguous dual pronunciations. Specifically:

- The character 'x' lacks a direct corresponding phoneme in Persian
- The characters 'c', 'q', 'u', and 'w' create ambiguous mappings as they share phonetic realizations with 'k', 'k', 'o', and 'v' respectively

We introduced specialized characters to represent distinct Persian phonemes to resolve these ambiguities. For instance, in the case of the Persian word 'خواب' (meaning 'sleep'), rather than using the conventional Pinglish transliteration 'khaab' or potentially ambiguous 'cAb', our system maps it to 'ķAb'. This systematic approach ensures that each Persian phoneme corresponds to a unique, unambiguous character in our intermediate representation. Figure 3 illustrates representative examples of input-output pairs from our grapheme-to-phoneme (G2P) model, demonstrating this systematic mapping.

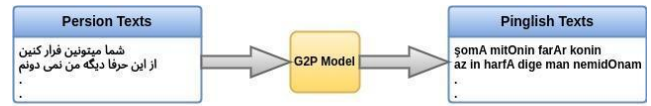

**Fig. 3.** Persian to intermediate language mapping examples using the proposed G2P model.

### 3.2 Homograph Disambiguation and Machine Transliteration Model

Our methodology encompasses homographs with multiple pronunciations disambiguation followed by machine translation. The training process utilized the previously constructed dataset, with particular attention to homograph representation.

### 3.3 Tokenization Strategy

We implemented a tokenization approach using one distinct byte-pair encoding (BPE) [23], specifically its adaptation for NLP tasks [24], tokenizers:

Custom BPE Tokenizer:

➔ Vocabulary size: 2372 tokens

➔ Minimum frequency threshold: 100

- ➔ Training corpus: 64,000 entries from the constructed dataset.
- ➔ Tendency towards subword tokenization with max-length smaller than four due to data characteristics and to accommodate limited vocabulary coverage

To design an effective tokenizer strategy, we interleaved Persian and Pinglish sentences with a maximum token length of less than 4, ensuring that the maximum subword length is limited to 3. This approach leads to a more consistent subword length distribution across Persian and Pinglish inputs. As a result, the model learns transliteration more effectively in the sequence-to-sequence framework and achieves better generalization performance.

We implemented a custom BPE tokenizer to address cross-linguistic elements in our corpus. This decision was motivated by the frequent occurrence of English words and characters within Persian texts. The custom approach enabled our system to handle these multilingual elements effectively, ensuring robust processing of mixed-language content while maintaining tokenization consistency across the entire corpus. This custom tokenization strategy allows our model to:

- Handle cross-linguistic elements effectively
- Optimize for different vocabulary distributions in source and target languages

### 3.4 Model Architecture Selection

The EncoderDecoder model [12],[13],[14] a highly efficient neural machine transliteration framework, was deployed with specific architectural parameters optimized for our task. Introduced initially as a fast and self-contained NMT framework, the model has proven effective for various transliteration tasks. For our NMT component, we implemented the Encoder-Decoder architecture trained from scratch. The architecture of the model is depicted in Figure 4.

The selection of EncoderDecoder was motivated by several critical considerations:

#### 3.4.1 Benchmark Performance

The architecture has demonstrated superior performance in MT benchmarks, making it particularly suitable for our G2P conversion task, which we approached as a transliteration problem.

#### 3.4.2 Computational Efficiency

A crucial factor in our selection was the model's inference time optimization, which is essential for text-to-speech (TTS) applications. EncoderDecoder's implementation with a small model embedding size significantly reduces inference latency, making it particularly suitable for real-time TTS systems.

#### 3.4.3 Contextual Understanding

The EncoderDecoder architecture provides robust contextual understanding, which is crucial for our task. This architectural feature enables:

- Comprehensive processing of input sequences
- Effective capture of long-range dependencies
- Enhanced semantic preservation during translation

The EncoderDecoder framework is particularly advantageous for our use case, as it facilitates the complex mapping between Persian graphemes and their corresponding phonemic representations while maintaining contextual integrity throughout the conversion process.

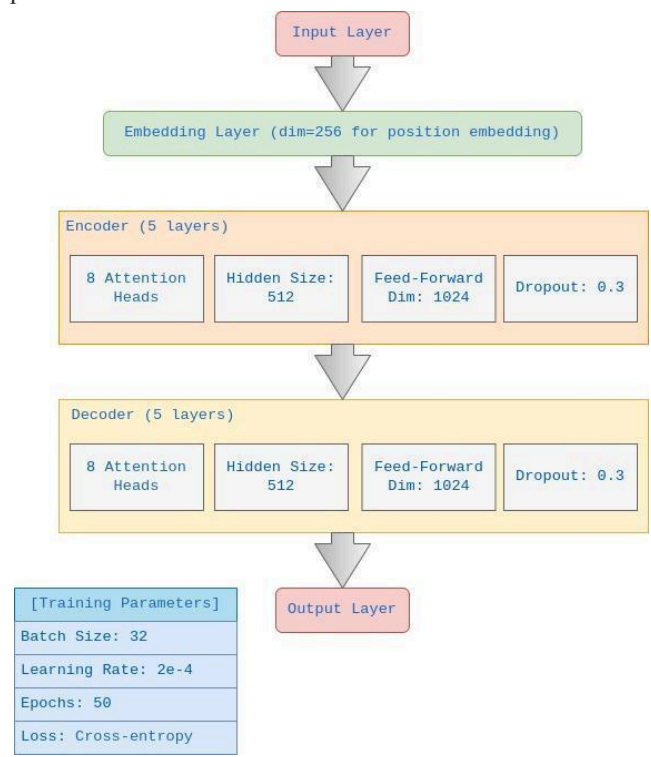

**Fig.4.** Schematic overview of the proposed architecture for the G2P model.

## 4. Experimental Setup

Our experimental framework utilized the dataset described in the previous section, implementing a strategic split and augmentation approach. From the initial 64,000 entries, we allocated 1,000 samples each for validation and test evaluation, with the remaining 62,000 samples designated for training data augmentation. Through sentence merging and strategic splitting at non-Ezafe positions, we expanded our training dataset to 195,000 samples, maintaining linguistic integrity throughout the augmentation process.

The EncoderDecoder model was implemented with specific architectural parameters optimized for our task. The encoder and decoder comprised 5 layers with 8 attention heads and a feed-forward dimension of 1,024. We maintained a hidden size of 512 across both components, utilizing a position embedding dimension of 256. The model was trained with a dropout rate of 0.3, a batch size of 32, and a learning rate of 2e-4 over 50 epochs. Our training protocol incorporated cross-entropy loss for evaluation and implemented an early stopping mechanism to prevent overfitting. For evaluation, we employed the BLEU metric [15] and PER [17], which are computed as:

$$BLEU = BP \times exp\left(\sum_{i=1}^{n} w_i \log p_i\right) \quad (1)$$
$$PER = (I + Del + S) / Total\ N \quad (2)$$

Where $BP$ denotes the brevity penalty, $w_i$ represents uniform weights, and $p_i$ indicates the modified $n$-gram precision. In Equation (2), $I$ denotes inserted phonemes, $Del$ indicates deleted phonemes, $S$ represents substituted phonemes, and $Total\ N$ refers to the total number of phonemes in the reference. The model achieved a BLEU score of 94.6 and a PER of 0.0196, demonstrating exceptional performance in the G2P conversion task. This high score validates the effectiveness of our architectural choices and training approach in capturing the nuances of Persian phoneme conversion.

## 5. Results and Analysis

To evaluate the effectiveness of our proposed model, we conducted extensive experiments on both our custom dataset and the SentenceBench dataset [3]. Table 2 presents the performance of our model in detecting Ezafe constructions over 50 epochs on our test set, demonstrating its ability to accurately infer implicit phonetic markers that are not explicitly represented in Persian script. The high detection accuracy highlights the model's capability to address one of the key phonological challenges in Persian G2P conversion. Table 3 reports the performance of our model in identifying homographs and presents the overall BLEU score and PER, which reflects the general quality of the model's predictions. The results indicate that our approach effectively resolves homograph ambiguity by leveraging contextual cues, contributing to improved phoneme prediction accuracy. Finally, Table 4 compares our model against an alternative approach on the SentenceBench dataset [1]. The results demonstrate the superiority of our model, as it achieves higher performance across key evaluation metrics. Notably, our model achieves higher homograph disambiguation accuracy on SentenceBench. However, its overall performance on this dataset is slightly lower than on our test set. This discrepancy can be attributed to the inconsistent statistical distribution of sentences containing homographs between the two datasets. This improvement underscores the effectiveness of our intermediate language representation and neural machine transliteration framework in handling complex phonological challenges in Persian.

**Table 2.** Performance metrics for Ezafe construction detection on the test set, showing the model's accuracy in identifying implicit phonetic markers in Persian text.

| Ezafe Performance Evaluation | |
|---|---|
| Precision | 0.99 |
| Recall | 0.99 |
| F1 score | 0.99 |

**Table 3.** Performance of the proposed model on homograph disambiguation and overall prediction quality (BLEU and PER scores).

| Our test set | |
|---|---|
| Homograph Acc. % | 84.00 |
| PER % | 1.96 |
| BLEU | 94.60 |

**Table 4.** Comparative performance of the proposed and best models on the SentenceBench datasets [1] across key evaluation metrics.

| | SentenceBench dataset[3] | |
|---|---|---|
| Ezafe F1 % | Sentence Level G2P[1] | OUR G2P |
| | 93.03 | **99.0** |
| Homograph Acc. % | Sentence Level G2P[1] | OUR G2P |
| | 78.50 | **82.60** |
| PER % | Sentence Level G2P[1] | OUR G2P |
| | 5.80 | **2.99** |

## 6. Conclusion

This paper presents a novel approach to Persian grapheme-to-phoneme conversion by developing an intermediate language system that effectively addresses the challenges of homographs with multiple pronunciations disambiguation and Ezafe detection. Our research makes several significant contributions to the field of Persian language processing. First, we introduced a robust intermediate language framework that bridges the gap between Persian text and phonemic representation, achieving a remarkable BLEU score of 94.6 and PER of 0.0196. This demonstrates the effectiveness of our approach in handling the complexities of Persian phonology.

The custom tokenization strategy proved particularly effective in handling cross-linguistic elements. Our methodology's success is further evidenced by our comprehensive dataset comprising 64,000 entries and expanded to 195,000 samples through strategic augmentation while maintaining linguistic integrity. Implementing the EncoderDecoder architecture, optimized with specific parameters for our use case, demonstrated both computational efficiency and superior performance in contextual understanding.

One of our most significant achievements is developing a cost-effective solution that maintains high accuracy while reducing the computational resources typically required for such tasks. This makes our approach particularly valuable for real-world applications and resource-constrained environments. The system's ability to handle both formal and informal Persian text, along with its robust homograph disambiguation capabilities, represents a substantial advancement in Persian language processing technology also the findings of this research are generalizable to other languages with complex linguistic structures involving homographs, including Chinese and Arabic.

## 7. Future work

Several promising research directions emerge from this work. The primary focus will be extending the intermediate language framework to morphologically rich languages, investigating language-specific adaptations for phonological complexities, and developing universal phonological mapping strategies. We plan to explore advanced contextual processing by implementing transformer-based architectures for enhanced homographs with multiple pronunciations disambiguation and the integration of semantic-aware processing modules. A significant avenue for investigation lies in developing multilingual MT models, focusing on handling multiple source languages and exploring cross-lingual transfer learning capabilities. Furthermore, we aim to enhance our dataset through large-scale corpus utilization and automated data augmentation techniques, enabling more robust and comprehensive training data. These research directions aim to improve the framework's robustness and generalizability while maintaining computational efficiency and accuracy in phoneme conversion tasks. The insights gained from these future investigations will contribute to the broader computational linguistics and speech-processing field, particularly in resource-constrained environments.